\documentclass[runningheads]{llncs}
\usepackage{graphicx}
\usepackage{caption}
\usepackage{subcaption}
\begin{document}
\title{Detecting Mitosis against Domain Shift using a Fused Detector and Deep Ensemble Classification Model for MIDOG Challenge}
\titlerunning{Detecting Mitosis against Domain Shift}
%
\author{Jingtang Liang\inst{1} \and
Cheng Wang\inst{1} \and Yujie Cheng\inst{2} \and Zheng Wang\inst{2} \and Fang Wang\inst{2} \and Liyu Huang\inst{1} \and Zhibin Yu\inst{2} \and Yubo Wang\inst{1}
}
\authorrunning{Liang et al.}
%
\institute{School of Life Science and Technology, Xidian University, Xi'an, Shannxi, China \and
College of Electrical Engineering, Ocean University of China, Qingdao, Shandong, China\\
\email{huangly@mail.xidian.edu.cn, yuzhibin@ouc.edu.cn, ybwang@xidian.edu.cn}}
\maketitle              
\begin{abstract}
Mitotic figure count is an important marker of tumor proliferation and has been shown to associated with patients prognosis. Deep learning based mitotic figure detection methods have been utilized to automatically locate the cell in mitosis using hematoxylin \& eosin (H\&E) stained images. However, the model performance deteriorates due to the large variation of color tone and intensity in H\&E images. In this work, we proposed a two stage mitotic figure detection framework by fusing a detector and a deep ensemble classification model. To alleviate the impact of color variation in H\&E images, we utilize both stain normalization and data augmentation, aiding model to learn color irrelevant features. The proposed model obtains a F1 score of 0.7550 on the preliminary testing set released by the MIDOG challenge.

\end{abstract}
\section{Introduction}

Tumor proliferation obtained form hematoxylin \& eosin stained (H\&E) histopathological images provides valuable information regarding the patient prognosis and treatment planning, especially in breast cancers \cite{VanDiest2004}. Mitotic activity of tumor cells observed in high power field view is an epiphenomenon of the cell proliferation, is therefore selected to quantify the tumor proliferation and has been shown to associated with the patients' prognosis \cite{Medri2003}.

In pathology labs, experienced pathologists count the cell under active mitosis phase in 10 consecutive high-power field views to calculate the mitotic index. This practice is time-consuming and prone to human error inducing high inter-observer disagreement \cite{Theissig1990}. Recently, owning to its superior performance in computer vision tasks, deep learning based methods have been extensively studied for automatically identifying tumor cells in mitosis using whole-slide images \cite{Chen2016,Li2018,Veta2015}.

The large variability observed in H\&E stained pathological images still impeded the application of automatic mitosis detection in clinical settings. The variability in H\&E images may arise due to the difference in the scanner to obtain WSI, or the procedure employed by different labs for tissue preparation and staining \cite{Aubreville2021,Howard2021}. In fact, it was recently shown that the WSI itself contains enough information to differentiate where it has been collected \cite{Howard2021}. Hence, deep mitosis detection algorithms that can generalize well across different scanners or from different institutions are desired.  

Despite the source of variation in the H\&E images, it mostly manifests as large variation in color tone among different H\&E images. Hence, earlier attempts primarily focused on unifying the color space by utilizing color normalization techniques \cite{Macenko2009}. Among them, the Macenko stain normalization has been employed in the prepossessing pipeline when training deep learning models \cite{AzevedoTosta2019} . Furthermore, the model can also acquire ability to generalize to out-of-distribution data by using data augmentation or using specially designed structure, such as domain-adversarial training \cite{Zhou2021a}. 

In this work, we presented our approach to the MIDOG challenge \cite{MarcAubreville2021}. Inspired by earlier works \cite{Li2018,Chen2016,Sohail2021}, we constructed a two stages mitosis detection model by using the detectorRS \cite{Qiao2020} as the base model to coarsely identify the mitosis figure in the images. The results of detector model is later refined by a deep ensemble classification model to illuminate false positives and improve the overall performance. To address the domain shift problem, we employed both stain normalization and data augmentation focusing on inducing color variation. Our results suggested two-stages model equipped with both stain normalization and data augmentation can be an potential solution to address the domain shift in detecting mitosis figures in H\&E images.

\section{Data-set}

The data-set was provided by the MIDOG challenge \cite{MarcAubreville2021}. In brief, all images were obtained from human breast cancer tissue samples after routine Hematoxylin \& Eosin staining.
The Training set consists of 200 H\&E images obtained from four different scanners, including Hamamatsu XR nanozoomer 2.0, Hamamatsu S360 (0.5 NA), Aperio ScanScope CS2, and Leica GT450. Each scanner provided 50 H\&E images. Mitotic figures were annotated for the first three scanners. In total, annotation of 1721 and mitotic figures and 2714 non-mitotic figures (hard negative cases) were provided. To train our model, we randomly selected 5 images from each scanner with annotation as the validation set. The rest of training images were used to optimize the model. 

The preliminary test set released by the the MIDOG challenge consisted of 20 WSIs from four scanners, in which two scanners were part of the training set and the remaining two scanners are unknown.

\section{Proposed Model}

\begin{figure}[!ht]
\centering
\includegraphics[height= 6cm]{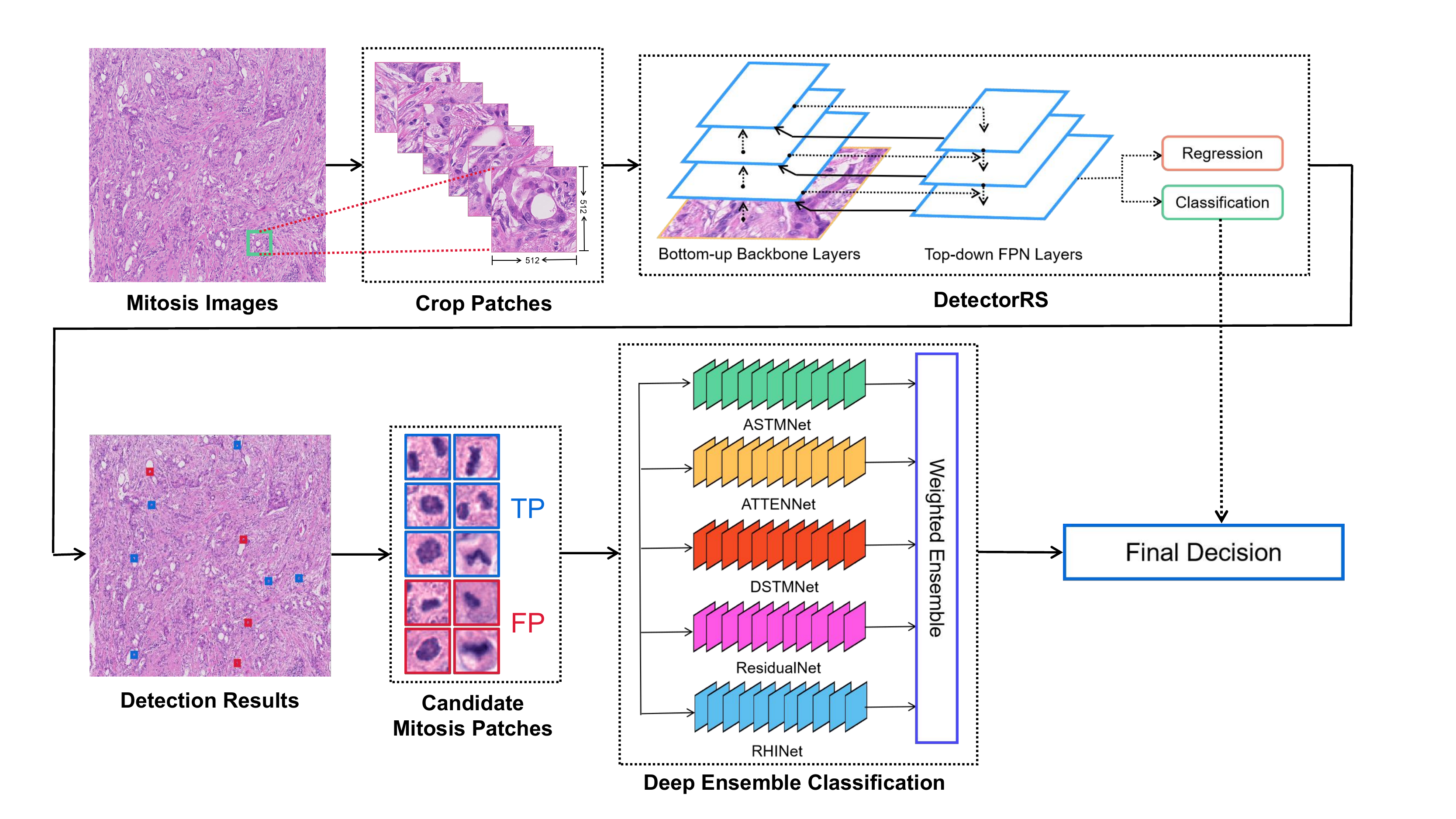}
\caption{Pipeline of the proposed mitotic figure detection framework}
\label{fig:1}
\end{figure}

The proposed model is shown in Fig.\ref{fig:1}. Our whole H\&E image processing pipeline consisted of five steps. Firstly, we cropped the original training images into patches of the size 512$\times$512 pixels, centered at the ground truth mitotic figures and hard negative cases. For each annotated cases, we randomly shifted the center of each patch within the range of $\pm$205 pixels. Then, a detectorRS model \cite{Qiao2020} was trained to identify the location of mitotic figures using a bounding box with a size of 50$\times$50 pixels. In the training phase, all training images were normalized with respect to the first images of the first scanner (\textit{001.tiff}) by using Macenko stain normalization. Then, we augmented the training patches by using random rotation, elastic deformation, scaling, Gaussian blur and a brightness and contrast enhancement. The detector was trained by using SGD with a learning rate of $0.02$ for $12$ epochs. Once the detector was trained, we employed the trained model to the whole training images to identify all suspected mitotic figures. To be noted, the model was trained using patches containing annotations of ground truth mitotic and hard negative cases, whereas the trained model scanned through the training images can produce many previously un-annotated false mitotic figures. This observation also motivated us to employed a second stage classifier to refine the results produced by the detector. 

The overall structure of deep ensemble model consists of five convolution networks, adopted from \cite{Sohail2021}. The input to the deep ensemble model was the suspected mitotic figures found by the trained detector on the training images with a classification threshold of $0.3$. The positive cases for training the classification model consisted of all samples with ground truth mitotic figures. The negative cases were the false positive cases identified by the detector on the whole training images and the hard negative cases. Training samples for the classification model were construed by shifting the center of obtained patches from detector within the range of [- 5, 5] pixels. To balance the training samples between positive and negative cases, we adjusted the number of times of applying offset to balance the number of cases in positive and negative classes. Finally, the samples were resized to 120$\times$120 and fed to the deep ensemble model. To overcome the domain shift caused by different scanner, we heavily utilized online augmentation methods that can induce color variation to increase the diversity of the training samples. The augmentation employed were horizontal and vertical flipping, random clipping and color jitter augmentation with luminance, contrast, hue and saturation disturbance intensity. Each individual model was optimized using AdamW with a learning rate of $2\times10^{-4}$ and was trained for 100 epochs. The optimal weights for each individual convolution network was selected based on their performance on the validation set. The output of the ensemble model was the weighted sum of soft-max score produced by each convolution networks. The final decision of the proposed two-stage mitotic figure detection was obtained by combing the classification score obtained from both detector and deep ensemble model as,
\begin{equation}
    S_{final} = \alpha*S_{DE} + (1-\alpha)*S_{Dect}
\end{equation}
where $\alpha \in [0,1]$ is the weights to balance the decision made by the detector and the deep ensemble model and optimized on the validation set, $ S_{final}$ is the final score to produce the final decision, $S_{DE}$ and $S_{Dect}$ are the classification score for the deep ensemble modular and detection modular, respectively. 

\section{Results}

We first tested the performance of detection modular on the validation set. The results of F1 score, precision and recall were given in Table.\ref{tab:1}. It can be observed that detector alone was able to retrieve almost 80\% of mitosis figures. In the meantime, it also produced many false positives resulted in a inferior precision score and a significantly degraded F1-score. The ability to refine the results obtained from the detector by the ensemble classification model weighted by different $\alpha$ was shown in Fig.\ref{fig:1}. It can be observed that by varying the value of $\alpha \in [0,1]$, the optimal performance on F1 score was found when $\alpha = 0.9$. The obtained model obtained a F1 score of 0.7550 on the preliminary testing set. 

\begin{table}[!ht]
    \centering
\begin{tabular}{|l|l|p{4cm}|p{4cm}|}
\hline
          & Detector Only & Proposed model on validation set & Proposed model on preliminary testing set  \\ \hline
F1-Score  & 0.5909      & 0.7128                                    & 0.7550                                             \\ \hline
Precision & 0.4719      & 0.7270            & 0.7238                                            \\ \hline
Recall    & 0.7904      & 0.6993         & 0.7892
 \\ \hline
\end{tabular}
    \caption{Model performance on validation set and preliminary testing set }
    \label{tab:1}
\end{table}

\begin{figure}[!ht]
\centering
\includegraphics[height= 6cm]{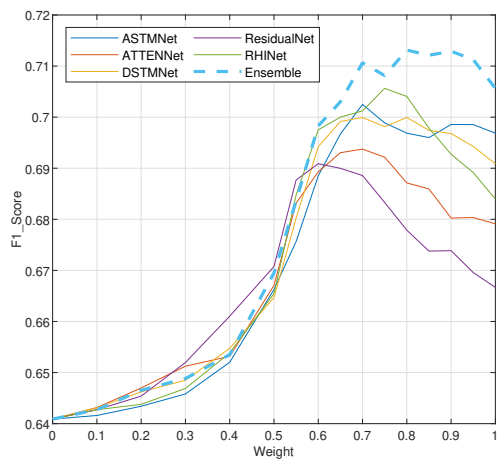}
\caption{Performance of individual and ensemble classification model on the validation set}
\label{fig:1}
\end{figure}


\section{Conclusion}

In conclusion, we presented a fused detector and deep ensemble classification model with image preprocessed by stain normalization and heavy data augmentation to address the domain shift problem for mitosis figure detection. Experiment results showed that the fused model performs reasonably well on the preliminary testing set released by the MIDOG challenge.

%
%
%
\bibliographystyle{splncs04}
%

\end{document}